\documentclass{article}

\usepackage[preprint]{neurips_2019}

\usepackage[utf8]{inputenc} 
\usepackage[T1]{fontenc}    
\usepackage{booktabs}       
\usepackage{amsfonts}       
\usepackage{nicefrac}       
\usepackage{microtype}      
\usepackage{graphicx}
\usepackage{amsmath}
\usepackage{caption}
\usepackage[table,xcdraw]{xcolor}
\usepackage{hyperref} 

\title{PidginUNMT: Unsupervised Neural Machine Translation from West African Pidgin to English}

\author{%
  Kelechi Ogueji\\
  InstaDeep\\
  \texttt{k.ogueji@instadeep.com} \\
 \And
   Orevaoghene Ahia\\
   InstaDeep \\
   \texttt{o.ahia@instadeep.com} \\
}
\begin{document}

\maketitle
\begin{abstract}
  Over 800 languages are spoken across West Africa. Despite the obvious diversity among people who speak these languages, one language significantly unifies them all - West African Pidgin English. There are at least 80 million speakers of West African Pidgin English. However, there is no known natural language processing (NLP) work on this language. In this work, we perform the first NLP work on the most popular variant of the language, providing three major contributions. First, the provision of a Pidgin corpus of over 56000 sentences, which is the largest we know of. Secondly, the training of the first ever cross-lingual embedding between Pidgin and English. This aligned embedding will be helpful in the performance of various downstream tasks between English and Pidgin. Thirdly, the training of an Unsupervised Neural Machine Translation model between Pidgin and English which achieves BLEU scores of 7.93 from Pidgin to English, and 5.18 from English to Pidgin. In all, this work greatly reduces the barrier of entry for future NLP works on West African Pidgin English.
\end{abstract}

\section{Introduction}
A lot of natural language processing (NLP) work has been done on the major languages in the world. However, little to no work has been done on the over 1800 African languages. The little work that has been done is on the major languages like Afrikaans, Zulu, Yoruba, Igbo, Hausa and Swahili. Pidgin English is one of the the most widely spoken languages in West Africa with 75 million speakers estimated in Nigeria as at 2016, and over 5 million speakers estimated in Ghana. The language originated from the Atlantic slave trade in the late 17th and 18th Centuries, where it was used by British slave merchants to communicate with the local African traders. It then spread across other West African regions because of its use as a trade language among regions who spoke different languages [1]. Even though different countries have different variants of Pidgin English, the language is fairly uniform across the continent. The variant of West African Pidgin English used in this work is the Nigerian Pidgin English (hereafter referred to as Pidgin), which has the highest population of speakers. 

This research work is the first - that we know of - that tackles a West African Pidgin English NLP problem.

In summary, this paper makes the following main contributions:

\begin{itemize}
    \item We provide the first Pidgin corpus containing 56,695 sentences and 32,925 unique words.
    \item We train cross-lingual word vectors between Pidgin and English, achieving a translation retrieval accuracy of of 0.1282 compared to the random baseline of 0.009.
    \item We train the first ever machine translation model between pidgin and English - an unsupervised neural machine translation model - achieving a BLEU score of 7.93 from Pidgin to English and 5.18 from English to Pidgin
\end{itemize}
\section{Related Work}
\paragraph{West African and Nigerian Pidgin English:} As stated earlier, there is no known NLP work on Pidgin. However, there has been a lot of linguistic work on the language, such as understanding its phonology and morphology [2]. Other works have studied Nigerian Pidgin, the most popular of all West African Pidgin variants [3, 4]. Another interesting work is a comparison of the variants of West African Pidgin English, which found that they are very similar [5]. 
\parskip -5pt
\paragraph{Cross-Lingual Embedding:}
There have been many successful methods for learning distributed representations of words [6,7,8], and these methods have greatly advanced NLP. However, these methods can only be used to learn representations for one language. 
\parskip -5pt
A lot of research has been done on learning representations that can be used across multiple languages. One such method [9] relied on the observation that continuous word embedding spaces possess similar properties across languages. They then learn a linear mapping between from one language space to another using a bilingual lexicon. Several other studies have further developed such supervised methods [10,11,12].  
\parskip -5pt
Some work has also been done on reducing the amount of supervision needed to learn cross-lingual embedding [13, 14]. Another work has been done on learning purely unsupervised cross-lingual embedding with unaligned monolingual corpora of each language. This method generates synthetic bilingual dictionary between a source and target language after using adversarial training to learn an initial mapping between both languages [14]. 
\parskip -5pt
There are several other cross-lingual embedding learning methods, and [16] does a good job of reviewing them.

\paragraph{Unsupervised Neural Machine Translation:}
The absence of parallel corpus for low-resourced languages has led to a dearth of machine translation models for these languages. For many of these languages, however, there exist monolingual corpus. Some works have been done on developing unsupervised machine translation models that use only these monolingual corpus, starting with [17] which treats non-parallel machine translation as a deciphering task. More recent works such as [18, 19, 20] have relied on three major principles - model initialization, language modelling and iterative back-translation - and have been largely successful. 
\parskip 2pt

\section{Study Methodology}

\subsection{Dataset}
In many low-resourced languages, there exists, at least, a wikipedia corpus. However, this is not the case for Pidgin English. The only available sources for large corpora for this language are news websites. 

We scraped a Pidgin news website [21] and obtained a corpus of 56695 sentences and 32925 unique words. Below are two examples of sentences in the dataset:

\begin{enumerate}
  \item \textit{dis one na one of di first songs wey commot dis year for nigeria but as dem release am, yawa dey.}
  \item \textit{dem say na serious gbege if dem catch anybody with biabia for inside di campus.}
\end{enumerate}

In general, the corpus was very messy and a lot of cleaning had to be done.

\subsection{Cross-lingual Embedding}
We trained cross-lingual embedding via monolingual mapping - where a linear mapping is learned between already trained monolingual word embedding [16]. We explored both supervised and unsupervised mapping methods.
\subsubsection {Pidgin Mono-lingual Embedding}
No prior word embedding exits for Pidgin, so we had to train the very first one. To do this, we used the Gensim library [22]. Given that Pidgin is a variant of English language and they share a lot of common words, we initialized the Pidgin embedding with an embedding of English. Also, Pidgin is a highly contextual language, so we needed to capture that in the word vectors. We initialized with Glove vectors [6] which gives a global context to word vectors. The fine-tuning on the Pidgin corpus was done with the continuous bag of words (CBOW) method [7] to give a more local context. The intuition was that the final word vectors would be able to capture both global and local contexts.

We trained 300-dimension vectors for 5 epochs with 8 negative samples, a window size of 5 and a batch size of 3000. 

\subsubsection {English Mono-lingual Embedding}
For this, we used pre-trained Glove vectors of dimension 300 [23]. 

\subsubsection {Unsupervised Cross-lingual Embedding}
For this, we used the MUSE library [24]. The method is based on [15] where a mapping is learned from the source to target language vector space using adversarial training and repetitive Procrustes refinement.
\subsubsection {Supervised Cross-lingual Embedding}
We explored two methods for learning supervised cross-lingual embedding. The first method was with the MUSE Library, which is based on iterative Procrustes alignment [15]. The second method was based on the retrieval criterion of [12], using the fastText alignment library [25].

In order to perform supervised alignment with these methods, we needed a bilingual dictionary with pairs of word translations between Pidgin and English. We scraped an online dictionary [26] and obtained 1097 word pairs. This dictionary was manually edited for errors and checked for translation credibility. 

\subsection{Unsupervised Neural Machine Translation}

The absence of a parallel corpus between at the start of this work meant we had to explore unsupervised machine translation. Training was done following the unsupervised neural machine translation model in [20] using the authors' UnsupervisedMT library [27].

We used a Transformer [28] with 10 attention heads. There are 4 encoder and 4 decoder layers with 3 encoder and decoder layers shared across both languages. The optimizer used for both the encoder and decoder is Adam [29] with a learning rate of 0.0003 and $\beta_1 = 0.5$. The batch size is 16 and we decode greedily. The discriminator used is a 3-layer feed-forward neural network with a hidden layer dimension of 128. We train it with the rmsprop [30] optimizer with a learning rate of 0.0005. 

At each training step, we perform the following:

\begin{enumerate}
  \item Discriminator training where we train a small neural network to predict the language of an encoded sentence. This ensures the decoder can translate regardless of the input source language.
  \item Denoising autoencoder training on each language (this is equivalent to training a language model as the model learns useful patterns for reconstruction and becomes immune to noisy input sentences). The encoder is trained to fool the discriminator such that latent representations of both source or target are indistinguishable.
  \item On-the-fly back translation such that a given sentence is translated with the current translation model (encoder and decoder), and we then attempt to reconstruct it from the translation.
\end{enumerate}

We trained for 8 epochs on a V100 (approximately 3 days). To select a model, we evaluated on a test parallel set of 2101 sentences from the JW300 [31] dataset pre-processed by the Masakhane group [32]. The model with the highest BLEU score [33] was selected as the best. 

\section{Results}

\subsection{Cross-lingual Embedding Task}
We evaluate using a word retrieval task which considers the problem of retrieving the correct translation of given a source word. We select the model that produces the highest translation retrieval precision (P@1) on a validation set of 108 word pairs. The baseline is the probability of randomly selecting the right translation word from the validation set. Table 1 shows the comparison of methods. 

\begin{table}[!h]
\centering
\begin{tabular}{|c|c|}
\hline
\textbf{Method} & \textbf{Precision at 1} \\ \hline
Random Selection Baseline      & 0.0093\\ \hline
Unsupervised Alignment [15]      & 0.0332\\ \hline
Supervised Procrustes Alignment [15]      & 0.0853\\ \hline
Supervised Alignment with a Retrieval Criterion [12]      & \textbf{0.1282}\\ \hline
\end{tabular}
\caption {Comparison of Methods for Cross-lingual Embedding Training }
\end{table}

\subsection{Unsupervised Neural Machine Translation Task}
After evaluating on the test set, the best model achieved a BLEU score of 7.93 from Pidgin to English and a BLEU score of 5.18 from English to Pidgin. Table 2 and Table 3 below shows some translation results by the model.

\textbf{Pidgin to English:}
\begin{table}[!h]
\centering
\begin{tabular}{ c|c }
\hline
Source      & dem dey really make us strong . \\ 
Reference      & they are a real source of encouragement . \\ 
Model Translation      & he 's really made us strong . \\ \hline

Source      & wetin we fit do to get better result when we dey preach for open place ?\\ 
Reference      & how can public witnessing prove to be effective ?\\ 
Model Translation      & what could we do to get better result when we preach in open place ?\\ \hline

\end{tabular}
\caption {Model Translation Results from Pidgin to English }
\end{table}

\textbf{English to Pidgin:}
\begin{table}[!h]
\centering
\begin{tabular}{ c|c }
\hline
Source      & what are most people today not aware of ?\\ 
Reference      & wetin many people today no know ? \\ 
Model Translation      & wetin most people are today no dey aware of \\
\hline
Source      & one student began coming to the kingdom hall .\\
Reference      & one of my student come start to come kingdom hall .\\ 
Model Translation      & one student wey begin dey come di kingdom hall .\\ 
\hline
\end{tabular}
\caption {Model Translation Results from English to Pidgin}
\end{table}

\section{Conclusion and Future Work}
The importance of West African Pidgin cannot be overstated, and natural language processing work on the language is bound to have an impact on tens of millions of people. 

In this work, we have presented the first known NLP work on the language. We began by providing the largest corpus of Pidgin that there is. We then trained the first ever Pidgin word vectors, aligning these with English word vectors to create cross-lingual embedding. Finally, we trained the first machine translation model with an unsupervised neural machine translation system. Future works include using byte pair encoding [34], instead of word vectors, testing this model with parallel data from other variants of West African Pidgin, qualitative evaluation with native speakers, and obtaining a supervised baseline with the JW300 dataset. 

The data, code and trained models have been made available \href{https://github.com/keleog/PidginUNMT}{https://github.com/keleog/PidginUNMT.} We hope this work spurs more natural language processing research on African languages. 

\section*{Acknowledgment}
We acknowledge our amazing colleagues at InstaDeep for useful discussions during this work, especially Karim Beguir, Tejumade Afonja, Lawrence Francis and George Igwegbe. We also thank Farouq Oyebiyi, Deepquest AI and AI Saturday's Lagos for support. Finally, special thanks to the Masakhane group for pre-processing the test data used to evaluate the model.

\section*{References}

\parskip 5pt
[1] The BBC (2016). Pidgin - West African lingua franca. https://www.bbc.com/news/world-africa-38000387

[2] Schneider, G. D. (1967). West African Pidgin-English--an Overview: Phonology-Morphology. Journal of English linguistics, 1(1), 49-56.

[3] Faraclas, N. (2002). Nigerian Pidgin. Routledge.

[4] Elugbe, B. O. \& Omamor, A. P. (1991). Nigerian Pidgin: Background and Prospects

[5] Peter, L., \& Wolf, H. G. (2007). A comparison of the varieties of West African Pidgin English. World Englishes, 26(1), 3-21.

[6] Pennington, J., Socher, R., \& Manning, C. (2014). Glove: Global vectors for word representation. In Proceedings of the 2014 conference on empirical methods in natural language processing (EMNLP) (pp. 1532-1543).

[7] Mikolov, T., Sutskever, I., Chen, K., Corrado, G. S., \& Dean, J. (2013). Distributed representations of words and phrases and their compositionality. In Advances in neural information processing systems (pp. 3111-3119).

[8] Bojanowski, P., Grave, E., Joulin, A., \& Mikolov, T. (2017). Enriching word vectors with subword information. Transactions of the Association for Computational Linguistics, 5, 135-146.

[9] Mikolov, T., Le, Q. V., \& Sutskever, I. (2013). Exploiting similarities among languages for machine translation. arXiv preprint arXiv:1309.4168.

[10] Artetxe, M., Labaka, G., \& Agirre, E. (2016, November). Learning principled bilingual mappings of word embeddings while preserving monolingual invariance. In Proceedings of the 2016 Conference on Empirical Methods in natural language processing (pp. 2289-2294).

[11] Ammar, W., Mulcaire, G., Tsvetkov, Y., Lample, G., Dyer, C., \& Smith, N. A. (2016). Massively multilingual word embeddings. arXiv preprint arXiv:1602.01925.

[12] Joulin, A., Bojanowski, P., Mikolov, T., Jégou, H., \& Grave, E. (2018). Loss in translation: Learning bilingual word mapping with a retrieval criterion. arXiv preprint arXiv:1804.07745.

[13] Smith, S. L., Turban, D. H., Hamblin, S., \& Hammerla, N. Y. (2017). Offline bilingual word vectors, orthogonal transformations and the inverted softmax. arXiv preprint arXiv:1702.03859.

[14] Artetxe, M., Labaka, G., \& Agirre, E. (2017, July). Learning bilingual word embeddings with (almost) no bilingual data. In Proceedings of the 55th Annual Meeting of the Association for Computational Linguistics (Volume 1: Long Papers) (pp. 451-462).

[15] Conneau, A., Lample, G., Ranzato, M. A., Denoyer, L., \& Jégou, H. (2017). Word translation without parallel data. arXiv preprint arXiv:1710.04087.

[16] Ruder, S., Vulić, I., \& Søgaard, A. (2017). A survey of cross-lingual word embedding models. arXiv preprint arXiv:1706.04902.

[17] Ravi, S., \& Knight, K. (2011, June). Deciphering foreign language. In Proceedings of the 49th Annual Meeting of the Association for Computational Linguistics: Human Language Technologies (pp. 12-21).

[18] Lample, G., Conneau, A., Denoyer, L., \& Ranzato, M. A. (2017). Unsupervised machine translation using monolingual corpora only. arXiv preprint arXiv:1711.00043.

[19] Artetxe, M., Labaka, G., Agirre, E., \& Cho, K. (2017). Unsupervised neural machine translation. arXiv preprint arXiv:1710.11041.

[20] Lample, G., Ott, M., Conneau, A., Denoyer, L., \& Ranzato, M. A. (2018). Phrase-based \& neural unsupervised machine translation. arXiv preprint arXiv:1804.07755.

[21] Pidgin Blog - https://Pidginblog.ng/

[22] Rehurek, R., \& Sojka, P. (2010). Software framework for topic modelling with large corpora. In In Proceedings of the LREC 2010 Workshop on New Challenges for NLP Frameworks.

[23] Pretrained Glove vectors (6B tokens, 400K vocab, uncased, 300d vectors) http://nlp.stanford.edu/data/glove.6B.zip

[24] MUSE: A library for Multilingual Unsupervised or Supervised word Embedding - https://github.com/facebookresearch/MUSE

[25] fastText alignment library - https://github.com/facebookresearch/fastText/tree/master/alignment/

[26] Naijalingo: Pidgin to English online dictionary - http://www.naijalingo.com/

[27] Phrase-Based \& Neural Unsupervised Machine Translation - https://github.com/facebookresearch/UnsupervisedMT

[28] Vaswani, A., Shazeer, N., Parmar, N., Uszkoreit, J., Jones, L., Gomez, A. N., Kaiser, L., \& Polosukhin, I. (2017). Attention is all you need. In Advances in neural information processing systems (pp. 5998-6008).

[29] Kingma, D. P., \& Ba, J. (2014). Adam: A method for stochastic optimization. arXiv preprint arXiv:1412.6980.

[30] Tieleman, Tijmen, \& Geoffrey Hinton. "Lecture 6.5-rmsprop: Divide the gradient by a running average of its recent magnitude." COURSERA: Neural networks for machine learning 4, no. 2 (2012): 26-31.

[31] Agić, Ž., \& Vulić, I. (2019). JW300: A wide-coverage parallel corpus for low-resource languages.

[32] Masakhane Group test set (https://www.masakhane.io/) - https://github.com/masakhane-io/masakhane/tree/master/jw300\_utils/test

[33] Papineni, K., Roukos, S., Ward, T., \& Zhu, W. J. (2002, July). BLEU: a method for automatic evaluation of machine translation. In Proceedings of the 40th annual meeting on association for computational linguistics (pp. 311-318). Association for Computational Linguistics.

[34] Sennrich, R., Haddow, B., \& Birch, A. (2015). Neural machine translation of rare words with subword units. arXiv preprint arXiv:1508.07909.

\end{document}